\documentclass{article}
    \usepackage[preprint]{neurips_2025}

\usepackage[utf8]{inputenc} % allow utf-8 input
\usepackage[T1]{fontenc}    % use 8-bit T1 fonts
\usepackage{url}            % simple URL typesetting
\usepackage{booktabs}       % professional-quality tables
\usepackage{amsfonts}       % blackboard math symbols
\usepackage{nicefrac}       % compact symbols for 1/2, etc.
\usepackage{microtype}      % microtypography
\usepackage{xcolor}         % colors
\usepackage{graphicx}
\usepackage{multirow} 
\usepackage{subcaption} 
\usepackage{makecell}
\usepackage{colortbl} 
\usepackage[misc]{ifsym}
\usepackage{amssymb}
\definecolor{cvprblue}{rgb}{0.21,0.49,0.74}
\usepackage[pagebackref,breaklinks,colorlinks,citecolor=cvprblue]{hyperref}

\newcommand{\methodname}{UniVLA}
\title{Unified Vision-Language-Action Model}

\author{%
  Yuqi Wang$^{1}\thanks{Work done during an internship at BAAI. $\textsuperscript{\Letter}$ Corresponding author}$ \quad
  Xinghang Li$^{2}$\quad
  Wenxuan Wang$^{1,2}$ \quad 
  Junbo Zhang$^{3}$ \quad
  Yingyan Li$^{1}$ \quad
  \\[1mm]
  \textbf{Yuntao Chen}$^4$ \quad
  \textbf{Xinlong Wang}$^2\textsuperscript{\Letter}$ \quad
  \textbf{Zhaoxiang Zhang}$^{1}$$\textsuperscript{\Letter}$
  \\[1mm]
  $^1$ CASIA \quad
  $^2$ BAAI  \quad
  $^3$ THU \quad
  $^4$ HKISI\\[1.5mm]
  Project page: \url{https://robertwyq.github.io/univla.github.io}
}

\begin{document}

\maketitle
\vspace{-8mm}
\begin{figure}[htbp]
    \centering
    \includegraphics[width=1.0\textwidth]{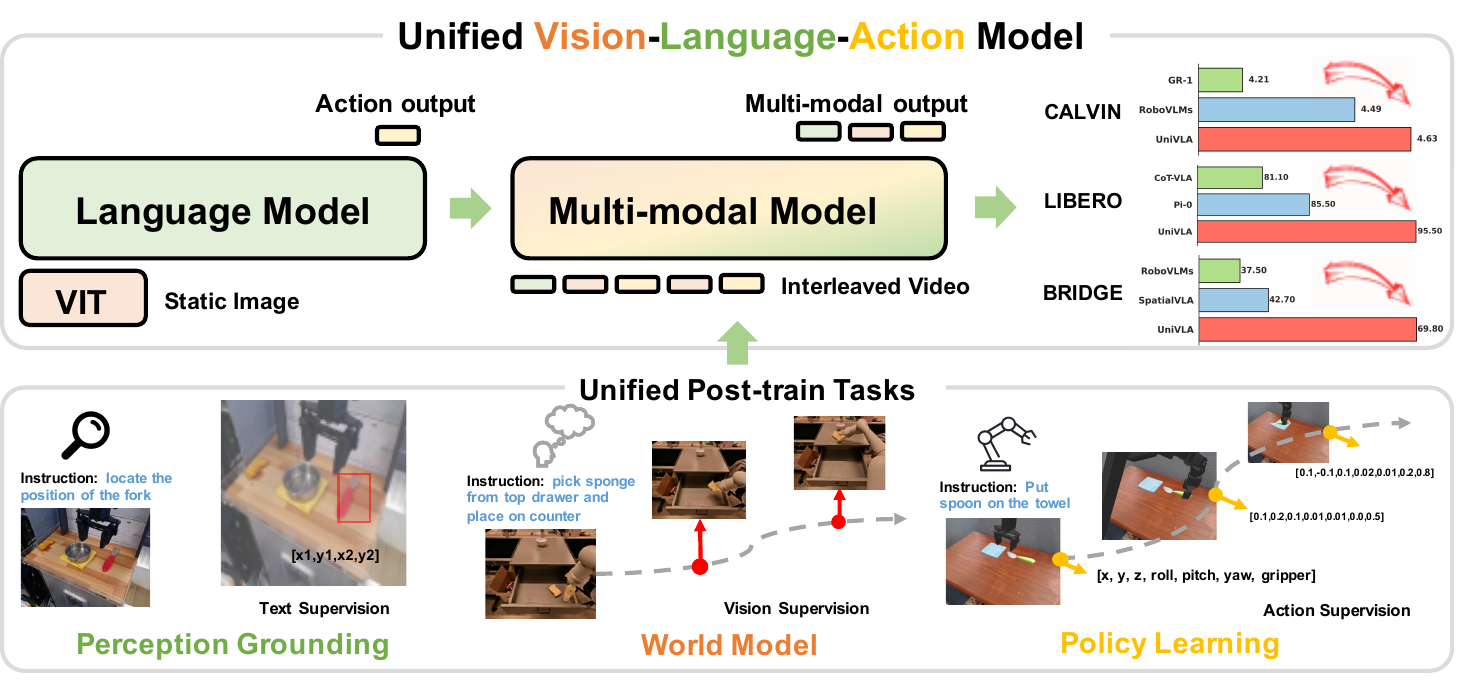}
    \caption{\textbf{We present \methodname{}, a unified vision-language-action model.} Unlike prior VLA approaches that typically rely on an extra vision encoder to extract image features and generate only action outputs, \methodname{} represents vision, language, and action as discrete tokens within a unified autoregressive framework. This unified modeling paradigm enables multi-modal outputs and supports a wide range of tasks—such as text-supervised perception grounding, vision-supervised world modeling, and action-supervised policy learning—within a single architecture. The unified token-based design further allows \methodname{} to effectively leverage large-scale multimodal data, particularly video, for scalable and generalizable learning. \methodname{} achieves new state-of-the-art results on CALVIN, LIBERO, and SimplerEnv-Bridge, significantly surpassing existing methods.}
    \label{fig:teaser}
\end{figure}
\begin{abstract}
  Vision-language-action models (VLAs) have garnered significant attention for their potential in advancing robotic manipulation.
However, previous approaches predominantly rely on the general comprehension capabilities of vision-language models (VLMs) to generate action signals, often overlooking the rich temporal and causal structure embedded in visual observations. In this paper, we present \methodname{}, a unified and native multimodal VLA model that autoregressively models vision, language, and action signals as discrete token sequences. This formulation enables flexible multimodal tasks learning, particularly from large-scale video data.
By incorporating world modeling during post-training, \methodname{} captures causal dynamics from videos, facilitating effective transfer to downstream policy learning—especially for long-horizon tasks.
Our approach sets new state-of-the-art results across several widely used simulation benchmarks, including CALVIN, LIBERO, and Simplenv-Bridge, significantly surpassing previous methods. 
For example, \methodname{} achieves 95.5\% average success rate on LIBERO benchmark, surpassing $\pi_0$-FAST's 85.5\%.
We further demonstrate its broad applicability on real-world ALOHA manipulation and autonomous driving.

\end{abstract}

\section{Introduction}
% background
Developing agents capable of perceiving, reasoning, and acting in the physical world has long been a central objective of artificial intelligence. Recent advances in vision-language-action (VLA) models~\cite{brohan2023rt, octo_2023, kim2024openvla, black2024pi_0}, grounded in the powerful generalization capabilities of vision-language models (VLMs)~\cite{peng2023kosmos, jaech2024openai,beyer2024paligemma, wang2024qwen2,guo2025deepseek}, have demonstrated impressive performance across a wide range of robotic manipulation tasks, and are increasingly being adapted to generalist humanoid robots~\cite{bjorck2025gr00t, ding2025humanoid} that demand broader embodied intelligence.
However, most existing VLA approaches~\cite{kim2024openvla,black2024pi_0} follow a language-centric paradigm: visual observations are first projected into a semantic space, and action policies are subsequently derived based on these representations. This late-fusion strategy, while beneficial for semantic understanding and generalization, limits the formation of deeply coupled cross-modal representations and impedes the learning of temporal and causal dependencies across the perception-action loop.
This raises a central question: \emph{Can vision, language, and action be jointly modeled within a unified representation space to facilitate tighter cross-modal integration and more effective policy learning?}

% challenges
While appealing in theory, unified modeling presents two key challenges. First, vision, language, and action are inherently heterogeneous modalities: vision comprises high-dimensional, continuous spatial signals; language conveys abstract, discrete semantics; and actions involve temporally ordered sequences with causal dependencies. 
Second, the perception-to-action pipeline is inherently dynamic and causal, yet existing VLA models~\cite{brohan2023rt, kim2024openvla, black2024pi_0} often adopt static, language-centric paradigms that merely learn the mapping from static image to action. These models fail to capture the dynamic nature of real-world interactions, thereby limiting their ability to leverage the rich temporal information from videos for training.

% our method
To address the above challenges, we introduce \methodname{}, a novel framework for unified vision–language–action learning.
As illustrated in Figure~\ref{fig:teaser}, we propose a unified framework that supports both \emph{multimodal} and \emph{multi-task} learning. At the modality level, vision, language, and action signals are all transformed into discrete tokens and modeled using a shared vocabulary. This unified token representation allows for joint learning across modalities, fostering deeper cross-modal understanding and integration.
Building upon the unified framework, we adopt an autoregressive, Markov chain-based sequence modeling approach, where observations and actions are interleaved. This structure naturally incorporates causal dependencies, enabling the model to reason over temporal dynamics rather than treating perception and action as isolated tasks. By integrating the world model paradigm during training, we leverage large-scale robotics videos for self-supervised learning, allowing the model to capture environment dynamics in a temporally consistent and causally grounded manner. Remarkably, we find that post-training with world models significantly enhances policy learning, particularly for long-horizon and out-of-distribution tasks.

% our findings
Experiments across multiple simulation benchmarks, including CALVIN~\cite{mees2022calvin}, LIBERO~\cite{liu2023libero}, and SimplerEnv~\cite{li2024evaluating}, demonstrating clear performance improvements over existing methods.
Our model incorporates world model learning during post-training, enabling it to effectively capture visual dynamics from large-scale videos. This strategy significantly enhances both data efficiency and training efficiency in downstream policy learning, and allows for rapid adaptation to novel robotic tasks. 
Beyond policy learning, we demonstrate the model’s multimodal output capabilities, including spatial reasoning and visual prediction, highlighting its versatility. 
Furthermore, we extend our approach to autonomous driving scenarios for broader applicability. These results underscore the potential of our unified VLA model as an alternative and promising direction for generalist embodied intelligence. 

Our contributions are summarized as follows:
\begin{itemize}
    \item We propose \methodname{}, the first unified vision–language–action (VLA) model that encodes vision, language, and action as discrete tokens within a shared vocabulary, jointly modeling them through autoregressive sequence learning. This approach offers a novel architecture alternative to the existing VLA paradigm, facilitating more integrated cross-modal modeling and enabling large-scale video-based training.
    \item Our unified sequence modeling framework supports a broad range of multimodal tasks. By investigating various post-training strategies, we demonstrate that world models can effectively learn temporal dynamics from video data, substantially enhancing performance and improving both data and training efficiency in downstream policy learning—particularly in long-horizon and out-of-distribution scenarios.
    \item Our model achieves state-of-the-art performance on several simulated benchmarks (CALVIN, LIBERO, and SimplerEnv-Bridge) and introduces an open-source VLA method supporting large-scale video training. We further explore its capabilities across various modalities, including spatial reasoning and video prediction, and demonstrate its effective transfer to driving scenarios, highlighting its potential for generalist embodied intelligence.
\end{itemize}
\section{Related Works}
\subsection{Vision-Language-Action Models}
Recent vision-language-action (VLA) models have demonstrated strong task performance across diverse robots and tasks~\cite{brohan2023rt,vuong2023open,driess2023palm, kim2024openvla,zhen20243d, cheang2024gr, black2024pi_0, zheng2024tracevla, liu2025hybridvla, kim2025fine, intelligence2025pi_}.
These models leverage pre-trained vision-language models (VLMs) to enhance understanding and generalization, further fine-tuned on large-scale robotic datasets for low-level control.
Currently, VLA models can be categorized into two paradigms based on their output space: \emph{pure action prediction} and \emph{visual-guided action prediction}.

\textbf{Pure action prediction.} 
Recent efforts have extended vision-language models (VLMs) to incorporate action modalities, enabling direct action prediction from visual and language inputs. A prominent example is RT-2\cite{brohan2023rt}, which learns from both internet-scale and robotic data to generate discrete actions autoregressively, showcasing strong generalization and semantic grounding. Building upon this, RT-H\cite{belkhale2024rt} introduces hierarchical actions to facilitate data sharing across tasks. OpenVLA\cite{kim2024openvla} scales this paradigm with a 7B-parameter open-source model trained on 970k real-world demonstrations spanning diverse datasets. To enhance spatial reasoning, SpatialVLA\cite{qu2025spatialvla} integrates spatial representations into the action modeling process.
Beyond architecture scaling, new action modeling techniques have also emerged. $\pi_0$~\cite{black2024pi_0} incorporates flow matching to improve action learning efficiency, while FAST ~\cite{pertsch2025fast} introduces a unified frequency-domain formulation for discretizing actions.

\textbf{Visual-guided action prediction.}
These studies leverage the power of visual pretraining, typically based on a policy-as-video formulation, by predicting future visual signals and subsequently decoding them into actions.
SuSIE~\cite{black2023zero} predicts key future frames and derives actions through inverse dynamics. UniPi~\cite{du2023learning} generates videos from text instructions, extracting actions from the frames. GR series~\cite{wu2023unleashing,cheang2024gr,li2025gr} leverages video pretraining for general policy learning. PAD~\cite{guo2024prediction} uses diffusion models to simultaneously learn future images and actions. LAPA~\cite{ye2024latent} proposes to learn latent actions between images with VQ-VAE from action-free internet-scale videos. Track2Act~\cite{bharadhwaj2024track2act} extracts point tracks from diverse web videos to guide the learning of interaction plans.

Both approaches have their strengths and weaknesses. The first, focused on action prediction, integrates well with Vision-Language Models (VLMs) but lacks spatial understanding and visual prediction capabilities. The second, incorporating visual generation, requires separating generative and action prediction models, limiting the full potential of VLMs. Our work unifies these approaches, combining video generation pretraining with the strengths of VLMs to propose a native multimodal model with significant future potential.

\subsection{World Models for Robotics}
World models~\cite{ha2018world, hafner2019dream, lecun2022path} have gained widespread attention for their ability to capture and reason about the dynamics of the physical world. 
They have emerged as a cornerstone in a range of domains, including interactive video generation~\cite{ bruce2024genie, che2024gamegen}, autonomous driving~\cite{hu2023gaia, wang2024driving, wang2024drivedreamer, gao2024vista}, and robotics~\cite{du2023learning, wu2023daydreamer, yang2023learning}.
Recent advances in robotics increasingly focus on general-purpose controllable video generation to simulate realistic and diverse robot-environment interactions. Visual Foresight~\cite{finn2017deep} leverages action-conditioned video prediction with model-predictive control, enabling robots to plan manipulation tasks by forecasting future visual observations. UniSim~\cite{yang2023learning} builds a “universal simulator” trained on diverse visual datasets, capable of visualizing the effects of both high-level instructions (e.g., “open the drawer”) and low-level controls in novel scenes. RoboDreamer~\cite{zhou2024robodreamer} learns a compositional world model by factorizing video generation, facilitating the synthesis of novel action sequences. DREMA~\cite{barcellona2024dream} replicates scene dynamics and structure by integrating Gaussian Splatting with physics simulation. VLP~\cite{du2023video} enables long-horizon visual planning by combining text-to-video generation with vision-language models as heuristic evaluators. DayDreamer~\cite{wu2023daydreamer} extends Dreamer~\cite{hafner2019learning} to real-world robotic platforms, while UVA~\cite{li2025unified} proposes a joint video-action latent space to decouple video and action generation, achieving high accuracy and efficiency in policy inference. AdaWorld~\cite{gao2025adaworld} extracts latent actions from videos in a self-supervised manner and builds an autoregressive world model conditioned on these latent actions.

\begin{figure}[tbp]
    \centering
    \includegraphics[width=1.0\linewidth]{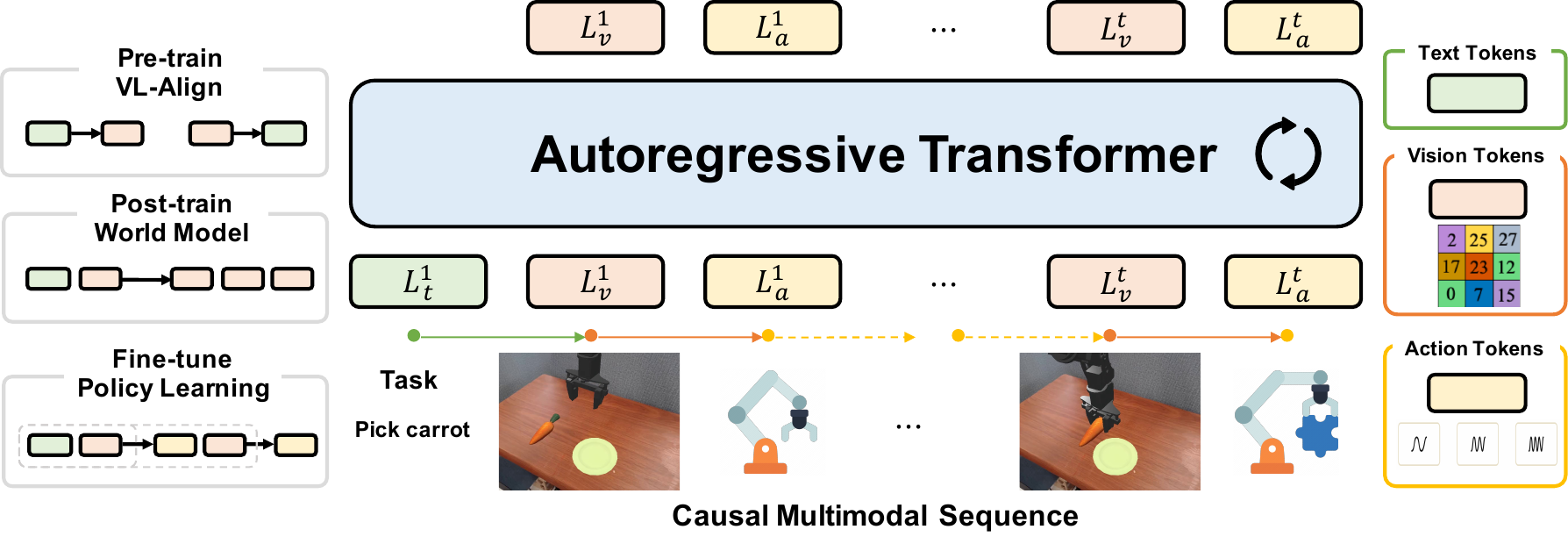}
    \caption{\textbf{Overview of the \methodname{} framework.} Our model unifies information from different modalities into a discrete interleaved sequence, which is modeled using an autoregressive Transformer. To enable unified modeling, images are discretized using vector-quantized (VQ) encoders, while actions are transformed into the frequency domain and discretized via Discrete Cosine Transform (DCT) encoding. This causal multimodal sequence naturally preserves the temporal dynamics and causality required for real-world tasks. The model builds upon a pretrained vision-language model and follows a two-stage training strategy: (1) a post-training phase that adopts world-model training on large-scale datasets without requiring actions, and (2) a fine-tuning phase that interleaves actions into the sequence, enabling policy learning on downstream tasks.}
    \label{fig:pipeline_univla}
    \vspace{-4mm}
\end{figure}

\section{Unified Vision-Language-Action Model}
In this section, we present the design of \methodname{}, as illustrated in Figure~\ref{fig:pipeline_univla}. Unlike previous VLA models~\cite{kim2024openvla, black2024pi_0} that rely on ViT~\cite{dosovitskiyimage} for image encoding, our approach adopts an encoder-free architecture, converting all modalities into discrete tokens and learning them autoregressively. The overall design is simple yet effective, demonstrating strong scalability.

Our \emph{unified paradigm} has two key aspects: first, it \emph{unifies the learning of multiple modalities}, integrating various modality tokens into a shared representation space and employing a transformer for autoregressive learning; Second, it \emph{unifies sequence modeling across tasks} through the natural interleaving of modalities, facilitating the seamless combination of tasks such as video generation, visual grounding, and action learning. In the following sections, we will introduce the method from the perspectives of \emph{Unified Multimodal Model} and \emph{Unified Multimodal Sequence Modeling}.

%---------------------------------------------------------------------------------------------------------
\subsection{Unified Multimodal Model}
% introduce model
As illustrated in Figure~\ref{fig:pipeline_univla}, our method unifies language, vision, and action modalities by converting each into discrete tokens and concatenating them into a single multimodal sequence \(L\). Specifically, \(L_t\), \(L_v\), and \(L_a\) denote the discrete token sequences for language, vision, and action, respectively, all drawn from a shared vocabulary. The superscript indicates the temporal step, with tokens interleaved across modalities to preserve temporal alignment.

For example, in the robotic manipulation task, a textual instruction is provided only at the beginning, followed by a naturally interleaved sequence of visual observations and actions. The language and vision tokenizers adopt the same design as Emu3~\cite{wang2024emu3}; visual observations are discretized using a VQ tokenizer~\cite{zheng2022movq}, while actions are encoded using FAST~\cite{pertsch2025fast}. To clearly demarcate modality boundaries, we employ special tokens—\texttt{boi} (begin of image), \texttt{eoi} (end of image), \texttt{boa} (begin of action), and \texttt{eoa} (end of action)—to encapsulate image and action tokens, respectively.

\vspace{-2mm}
\paragraph{Action Modeling}
We follow FAST~\cite{pertsch2025fast} and apply the Discrete Cosine Transform (DCT) to convert continuous action sequences into discrete action tokens. Specifically, given an action sequence within a time window, we define \( L_a \) at a given time step as a sequence of action tokens \( [T_1, \dots, T_n] \). The raw action sequence \( A_{1:H} = \{a_1, a_2, \dots, a_H\} \) spans a window of size \( H \), where each action \( a_t \) is a \( d \)-dimensional vector. The FAST action tokenizer encodes \( A_{1:H} \) into a discrete token sequence \( [T_1, \dots, T_n] \), with \( n \) tokens drawn from a vocabulary of size \( |V| \). Similar to natural language processing, action sequences can vary in token length, resulting in a variable-length ($n$) discrete representation.
\vspace{-2mm}
\paragraph{Training Objective}
Since all modality signals are transformed into discrete tokens, the training objective is simplified to a standard next-token prediction task using cross-entropy loss. To accommodate different task formats, we selectively include specific tokens in the loss computation, ensuring compatibility and flexibility across diverse tasks.
%---------------------------------------------------------------------------------------------------------
\subsection{Unified Multimodal Sequence Modeling}
% introduce task modeling
As shown in Figure~\ref{fig:pipeline_univla}, our multimodal sequence representation naturally captures the temporal dynamics and causal structure inherent in task execution. The embodied planning problem can be formulated as a Markov Decision Process (MDP), a general mathematical framework for decision-making in partially stochastic environments. For example, in the task of picking a carrot, the instruction and current observation inform the action; this action alters the environment, leading to a new observation that subsequently guides the next action. Building on this interleaved Markovian formulation, we unify a variety of tasks within a shared sequence modeling framework, and present the task-specific modeling strategies in the following.

\vspace{-2mm}
\paragraph{World Model (Post-train)}
Within the MDP framework, a world model aims to learn the dynamics of the environment by modeling the transition function \( P(\mathbf{s}_{t+1} | \mathbf{s}_t, \mathbf{a}_t) \). The learned world model enables agents to simulate future trajectories, plan actions, and reason about consequences without direct interaction with the environment. Specifically, in the context of robotic tasks, we treat the language instructiom as a general form of action. Given the current observation $L_v^1$ and the instruction $L_t^1$, the world model need to predict future visual content. In this setting, we use the loss computed solely from the vision tokens as the supervisory signal, enabling the model to generate visual predictions conditioned on the given instruction and observed state. Sequence $S_v$ formulation is as follows:
\begin{equation}
    S_v = \{L_t^1, L_v^1, L_v^2, ..., L_v^t\}
\end{equation}
\vspace{-2mm}
\paragraph{Policy Learning (Fine-tune)} 
Policy learning enables the agent to determine optimal actions based on both current observations and prior states, thereby effectively guiding task execution. In this setting, we employ a loss function computed solely from the action tokens. The sequence \( S_a \) representing the interactions over time is formulated as:
\begin{equation}
    S_a = \{L_t^1, L_v^1, L_a^1, L_v^2, L_a^2, \dots, L_v^t, L_a^t\}
\end{equation}

As illustrated in Figure~\ref{fig:pipeline_univla}, in this interleaved format, we adopt a two-stage training paradigm for robotic tasks. The model is initialized with a vision-language (VL) aligned checkpoint, endowing it with basic vision-language capabilities. The post-training stage leverages a world model objective to capture video dynamics, treating world modeling as a general visual learning task. Building upon the learned world model, the fine-tuning stage focuses on action learning to refine task-specific behaviors. We observe that incorporating the world model significantly enhances the efficiency and effectiveness of policy learning.

\section{Experiments}
\subsection{Dataset}
\paragraph{CALVIN.}
\emph{CALVIN}~\cite{mees2022calvin} is a simulated benchmark tailored for evaluating long-horizon, language-conditioned robotic manipulation. It comprises four simulated environments (A, B, C, and D), each containing demonstration trajectories collected via human teleoperation. The benchmark encompasses 34 distinct manipulation tasks with a total of 1,000 unique language instructions. Performance is measured by the average number of successfully completed sub-tasks within a sequence. Standard evaluation protocols include the \emph{ABC$\rightarrow$D} and \emph{ABCD$\rightarrow$D} settings, which test a model’s ability to generalize to unseen environments and compositions of long-horizon tasks.
\vspace{-2mm}
\paragraph{LIBERO.} 
The \emph{LIBERO} benchmark~\cite{liu2023libero} is a comprehensive suite for lifelong robotic manipulation, comprising four task suites with 10 tasks and 50 human demonstrations each. These suites are designed to evaluate different generalization abilities: \emph{LIBERO-Spatial} tests spatial reasoning by varying layouts with fixed objects; \emph{LIBERO-Object} assesses object-level generalization with varying objects in a fixed scene; \emph{LIBERO-Goal} targets goal-conditioned behavior by varying task goals; and \emph{LIBERO-Long} (\emph{LIBERO-10}) features long-horizon, compositional tasks with diverse objects, layouts, and goals, challenging temporal and compositional reasoning.
\vspace{-2mm}
\paragraph{SimplerEnv.} 
SimplerEnv~\cite{li2024evaluating} serves as a simulation benchmark designed to evaluate the transferability and generalization capabilities of models trained on real-world video data. It incorporates diverse manipulation setups across both the WidowX and Google Robot platforms, encompassing variations in lighting conditions, object textures, color distributions, and camera viewpoints.

% ----------------------------------------------------------------------------------------------
\subsection{Implementation Details}
The model adopts a purely autoregressive Transformer architecture with 8.5 billion parameters, identical to Emu3~\cite{wang2024emu3}. Images are tokenized using a VQ-based image encoder with a spatial compression factor of 8. For action encoding, we use the relative differences between consecutive frames. We first apply 1st and 99th percentile normalization, and then utilize the FAST tokenizer~\cite{pertsch2025fast}, which has a vocabulary size of 1024 and replaces the final 1024 token IDs of the language tokenizer.

\vspace{-2mm}
\paragraph{Post-training Stage.}
In the post-training stage, we leverage large-scale robot-centric video datasets to study the effects of various post-training strategies on downstream policy learning. The model is initialized with pre-trained weights from the first stage of Emu3~\cite{wang2024emu3}. We curate a total of 622K videos from existing robotics datasets (details provided in the appendix), and identify the world model as the most effective post-training approach. During training, supervision is applied solely on the vision tokens. The model is trained for 30K steps with a batch size of 64.

\vspace{-2mm}
\paragraph{Fine-tuning Stage.}
During fine-tuning, the model is initialized with weights from the post-training stage and trained using a two-frame interleaved vision-action sequence with an action chunk size of 10. A cosine annealing learning rate schedule is applied, starting at \(8 \times 10^{-5}\), and the loss is computed solely over action tokens.
For the CALVIN benchmark, RGB observations from both third-person (\(200 \times 200\)) and wrist-view (\(80 \times 80\)) cameras are used. Training is conducted on A100 GPUs with a batch size of 192 for 8k steps.
For the LIBERO benchmark, third-person and wrist-view RGB images (both at \(200 \times 200\)) are used to train a unified model with a batch size of 192 for 8k steps. A single model is evaluated across four task suites.
For the SimplerEnv benchmark, single-view RGB observations are used with input resized to \(256 \times 256\). Training is conducted on the Bridge-WidowX setup using a batch size of 128 for 20k steps, with an action chunk size of 5.

Additional implementation details on the post-training strategy, real-robot fine-tuning procedures, and autonomous driving experiments are provided in the appendix.
% ----------------------------------------------------------------------------------------------
\begin{table}[ht]
    \centering
    \caption{\textbf{Long-horizon robotic manipulation evaluation on the CALVIN benchmark.}}
      \resizebox{0.86\textwidth}{!}{
    \begin{tabular}{l l c c c c c c}
        \toprule
        \textbf{Method}  & \textbf{Task} & \multicolumn{5}{c}{\textbf{Tasks Completed in a Row}} & \textbf{Avg. Len $\uparrow$} \\
        \cmidrule(lr){3-7}
        &  & 1 & 2 & 3 & 4 & 5 & \\
        \midrule
        MCIL~\cite{lynch2020language} & ABCD$\rightarrow$D &0.373 & 0.027 & 0.002 & 0.000 & 0.000 & 0.40\\
        RT-1~\cite{brohan2022rt}  & ABCD$\rightarrow$D & 0.844 & 0.617 & 0.438 & 0.323 & 0.227 & 2.45 \\
        Robo-Flamingo~\cite{li2024vision} & ABCD$\rightarrow$D & 0.964 & 0.896 & 0.824 & 0.740 & 0.660 & 4.09 \\
        GR-1~\cite{wu2023unleashing}  & ABCD$\rightarrow$D & 0.949 & 0.896 & 0.844 & 0.789 & 0.731 & 4.21 \\
        UP-VLA~\cite{zhang2025up}  & ABCD$\rightarrow$D & 0.962 & 0.921 & 0.879 & 0.842 & 0.812 & 4.42 \\
        RoboVLMs~\cite{li2024towards} & ABCD$\rightarrow$D & 0.967 & 0.930 & 0.899 & 0.865 & 0.826 & 4.49 \\
        \rowcolor[gray]{0.9} \textbf{\methodname{}} & ABCD$\rightarrow$D & \textbf{0.985} & \textbf{0.961} & \textbf{0.931} & \textbf{0.899} & \textbf{0.851} & \textbf{4.63}\\
        \midrule
        MCIL~\cite{lynch2020language} & ABC$\rightarrow$D &0.304	&0.013&	0.002	&0.000	&0.000&	0.31\\
        Robo-Flamingo~\cite{li2024vision} & ABC$\rightarrow$D & 0.824 & 0.619 & 0.466 & 0.331 & 0.235 & 2.47 \\
        SuSIE~\cite{black2023zero}& ABC$\rightarrow$D& 0.870 &0.690 &0.490 &0.380 &0.260 &2.69\\
        GR-1~\cite{wu2023unleashing}  & ABC$\rightarrow$D & 0.854 & 0.712 &0.596	&0.497	&0.401	&3.06 \\
        UP-VLA~\cite{zhang2025up}  & ABC$\rightarrow$D & 0.928 &0.865& 0.815& 0.769& 0.699& 4.08 \\
        RoboVLMs~\cite{li2024towards} & ABC$\rightarrow$D &
        0.980 &0.936 &0.854& 0.778 &0.704 &4.25 \\
        Seer-Large~\cite{tian2024predictive} & ABC$\rightarrow$D &0.963 &0.916 &0.861 &0.803 &0.740 & 4.28\\
         \rowcolor[gray]{0.9} \textbf{\methodname{}}  & ABC$\rightarrow$D & \textbf{0.989} & \textbf{0.948} & \textbf{0.890} & \textbf{0.828} & \textbf{0.751} & \textbf{4.41}\\
        \bottomrule
    \end{tabular}}
        \label{tab:calvin_results}
\end{table}

\subsection{Main Results}
In this section, we evaluate our method on three simulation benchmarks: CALVIN (long-horizon tasks), LIBERO (diverse generalization), and SimplerEnv (real-to-sim manipulation). Our approach consistently achieves state-of-the-art performance across all settings.
\vspace{-2mm}
\paragraph{CALVIN Simulation Evaluation.}
Table~\ref{tab:calvin_results} presents the experimental results in the CALVIN benchmark. Our method achieves the highest performance on both the ABC$\rightarrow$D and ABCD$\rightarrow$D tasks, significantly outperforming previous approaches and demonstrating strong capabilities in multi-task learning and long-horizon planning.
\vspace{-2mm}
\paragraph{LIBERO Simulation Evaluation.}
Following~\cite{zhao2025cot}, we report the average success rate over 500 episodes for each task suite (Spatial, Object, Goal, Long). As shown in Table~\ref{tab:libero}, \methodname{} achieves the best overall performance across all LIBERO benchmark suites, with particularly significant gains on long-horizon tasks—improving the previous state of the art from 69.0\% to 94.0\%. Compared to $\pi_0$~\cite{pertsch2025fast}, our method demonstrates superior performance on long-horizon tasks.
\begin{table}[ht]
\centering
\caption{\textbf{Comparison of different methods on the LIBERO benchmark.}}
\resizebox{0.68\textwidth}{!}{
\begin{tabular}{lccccc}
\toprule
\textbf{Method} & \textbf{SPATIAL} & \textbf{OBJECT} & \textbf{GOAL} & \textbf{LONG} & \textbf{Average} \\
\midrule
DP*~\cite{chi2023diffusion}         & 78.3\% & 92.5\% & 68.3\% & 50.5\% & 72.4\% \\
Octo~\cite{team2024octo}      & 78.9\% & 85.7\% & 84.6\% & 51.1\% & 75.1\% \\
OpenVLA~\cite{kim2024openvla}& 84.9\% & 88.4\% & 79.2\% & 53.7\% & 76.5\% \\
SpatialVLA~\cite{qu2025spatialvla} & 88.2\% & 89.9\% & 78.6\% & 55.5\% & 78.1\% \\
CoT-VLA~\cite{zhao2025cot} & 87.5\% & 91.6\% & 87.6\% & 69.0\% & 81.1\%\\
$\pi_0$-FAST~\cite{pertsch2025fast} &\textbf{96.4\%}	& 96.8\%&	88.6\%&	60.2\%&	85.5\% \\
% \rowcolor[gray]{0.9} \textbf{\methodname{}} &\textbf{96.6\%} & \textbf{99.4\%} & \textbf{91.6\%} & \textbf{89.2\%} & \textbf{94.2\%}\\
\rowcolor[gray]{0.9} \textbf{\methodname{}} &95.4\% & \textbf{98.8\%} & \textbf{93.6\%} & \textbf{94.0\%} & \textbf{95.5\%}\\
\bottomrule
\end{tabular}}
\label{tab:libero}
\vspace{-3mm}
\end{table}

\vspace{-2mm}
\paragraph{SimplerEnv Simulation Evaluation.}
Table~\ref{tab:simplerenv_bridge} summarizes the performance across various manipulation policies on the Bridge-WidowX setup. Our approach demonstrates a significant improvement over prior methods, raising the average success rate from 42.7\% to 69.8\%. In particular, it shows marked improvements on previously difficult tasks, including stack block, put carrot and put spoon.
\begin{table}[htbp]
\centering
\caption{\textbf{Evaluation on SimplerEnv-WidowX across various manipulation tasks.}}
\label{tab:simplerenv_bridge}
\resizebox{\textwidth}{!}{
\begin{tabular}{l|cc|cc|cc|cc|c}
\toprule
\multirow{2}{*}{\textbf{Model}} 
& \multicolumn{2}{c|}{\textbf{Put Spoon on Towel}} 
& \multicolumn{2}{c|}{\textbf{Put Carrot on Plate}} 
& \multicolumn{2}{c|}{\textbf{Stack Green on Yellow Block}} 
& \multicolumn{2}{c|}{\textbf{Put Eggplant in Yellow Basket}} 
& \textbf{Overall} \\
\cmidrule(lr){2-3} \cmidrule(lr){4-5} \cmidrule(lr){6-7} \cmidrule(lr){8-9}
& \multicolumn{1}{c}{Grasp} & \multicolumn{1}{c|}{Success} 
& \multicolumn{1}{c}{Grasp} & \multicolumn{1}{c|}{Success} 
& \multicolumn{1}{c}{Grasp} & \multicolumn{1}{c|}{Success} 
& \multicolumn{1}{c}{Grasp} & \multicolumn{1}{c|}{Success} 
& \multicolumn{1}{c}{Success} \\
\midrule
RT-1-X~\cite{brohan2023rt}     
& 16.7\% & 0.0\%   
& 20.8\% & 4.2\%   
& 8.3\%  & 0.0\%   
& 0.0\%  & 0.0\%   
& 1.1\% \\
Octo-Base~\cite{octo_2023} 
& 34.7\% & 12.5\%  
& 52.8\% & 8.3\%   
& 31.9\% & 0.0\%   
& 66.7\% & 43.1\%  
& 16.0\% \\
Octo-Small~\cite{octo_2023}
& 77.8\% & 47.2\%  
& 27.8\% & 9.7\%   
& 40.3\% & 4.2\%   
& 87.5\% & 56.9\%  
& 29.5\% \\
OpenVLA~\cite{kim2024openvla}  
& 4.1\%  & 0.0\%   
& 33.3\% & 0.0\%   
& 12.5\% & 0.0\%   
& 8.3\%  & 4.1\%   
& 1.0\% \\
RoboVLMs~\cite{li2024towards}
& 70.8\% & 45.8\%  
& 33.3\% & 20.8\%  
& 54.2\% & 4.2\%  
& 91.7\% & 79.2%  
& 37.5\% \\
SpatialVLA~\cite{qu2025spatialvla} 
& 20.8\% & 16.7\% 
& 29.2\% & 25.0\% 
& 62.5\% & 29.2\% 
& 100\% & \textbf{100\%} 
& 42.7\% \\
% \rowcolor[gray]{0.9} \textbf{\methodname{}} 
% & \textbf{91.7\%} & \textbf{70.8\%} 
% & \textbf{66.7\%} & \textbf{58.3\%} 
% & \textbf{79.2\%} & \textbf{33.3\%} 
% & 95.8\% & 95.8\%  
% & \textbf{64.6\%} \\
\rowcolor[gray]{0.9} \textbf{\methodname{}} 
& \textbf{83.3\%} & \textbf{83.3\%} 
& \textbf{74.0\%} & \textbf{66.7\%} 
& \textbf{95.8\%} & \textbf{33.3\%} 
& \textbf{100.0\%} & 95.8\%  
& \textbf{69.8\%} \\
\bottomrule
\end{tabular}}
\vspace{-3mm}
\end{table}

% \input{tables/simplerenv_google}
% -----------------------------------------------------------------------
\subsection{In-Depth Analysis}
In this section, we provide an in-depth analysis within our unified framework, which may offer key insights for the design of future VLA models.
We first analyze how post-training enhances downstream policy learning in terms of both performance (Table~\ref{abl:post_train}) and training efficiency (Table~\ref{tab:post_efficiency}), highlighting the potential of world models as a general post-training strategy for robotics.
We then investigate that even without post-training stage, incorporating visual prediction loss (Table~\ref{tab:calvin_abl_visual}) and historical context (Table~\ref{tab:calvin_abl_history}) still contributes positively to policy learning. 
\vspace{-2mm}
\paragraph{Effectiveness of World Model Post-Training.}
Table~\ref{abl:post_train} investigates the effects of different post-training strategies on downstream policy learning across various simulation benchmarks. The results reveal that, due to inconsistencies in the action space across tasks, action-only learning exhibits low transferability, leading to a negative impact on performance. In contrast, most post-training approaches significantly enhance policy learning, highlighting the crucial role of visual learning in transferability.
Among these, the world model post-training approach yields the most substantial gains, enhancing both generalization and long-horizon planning capabilities.
A comparison with text-to-image (T2I) training emphasizes the importance of modeling temporal dynamics in video data, while contrasting with video-only training highlights the essential role of textual guidance in state transitions. Notably, this world model training requires no action annotations, enabling scalable learning from large-scale video data and providing a promising direction for future VLA research.
\definecolor{DeepGreen}{rgb}{0.0, 0.8, 0.0}

\begin{table}[tbp]
\centering
\caption{\textbf{Effectiveness of World Model Post-Training.} We compare different post-training strategies by fine-tuning only with action prediction on the downstream benchmarks.}
\resizebox{\linewidth}{!}{
\begin{tabular}{ccc|cc|cc}
\toprule
\multicolumn{3}{c|}{\textbf{Post-training Stage}} & \multicolumn{2}{c|}{\textbf{Generalization}} & \multicolumn{2}{c}{\textbf{Long-horizon}} \\
\textbf{Strategy} & \textbf{Sequence} & \textbf{Supervision} & \textbf{LIBERO} & \textbf{SimplerEnv-WidowX} & \textbf{LIBERO-Long} & \textbf{CALVIN} \\
\midrule
& & &  48.5 & 0.0 & 17.4 & 1.46 \\
action prediction & $T, I, A$ & action &43.9 (\textcolor{red}{-4.6}) & 0.0 & 10.6 (\textcolor{red}{-6.8})& 0.52(\textcolor{red}{-0.94})\\
text-to-image    & $T,I$ & vision & 69.8 (\textcolor{DeepGreen}{+21.3}) & 6.3 (\textcolor{DeepGreen}{+6.3}) & 55.8 (\textcolor{DeepGreen}{+38.4}) & 3.79 (\textcolor{DeepGreen}{+2.33}) \\
video prediction & $I_1,...,I_t$ & vision & 78.9 (\textcolor{DeepGreen}{+30.4})& 17.7 (\textcolor{DeepGreen}{+17.7}) & 80.8 (\textcolor{DeepGreen}{+63.4}) & 3.59 (\textcolor{DeepGreen}{+2.13}) \\
\rowcolor[gray]{0.9} world model      & $T,I_1,...,I_t$ & vision & \textbf{94.2} (\textcolor{DeepGreen}{+45.7}) & \textbf{64.6} (\textcolor{DeepGreen}{+64.6}) & \textbf{89.2} (\textcolor{DeepGreen}{+71.8}) & \textbf{4.61} (\textcolor{DeepGreen}{+3.15}) \\
\bottomrule
\end{tabular}}
\label{abl:post_train}
\vspace{-5mm}
\end{table}

\vspace{-2mm}
\paragraph{Data and Training Efficiency.}
Table~\ref{tab:post_efficiency} shows that post-training substantially enhances downstream policy learning efficiency. On the CALVIN benchmark (Table~\ref{tab:data_efficiency}), our method achieves higher success rates using only 10\% of the fine-tuning data, outperforming prior approaches such as GR-1~\cite{wu2023unleashing} and RoboVLMs~\cite{li2024towards}. In addition, Table~\ref{tab:train_efficiency} highlights improved training efficiency, as the model rapidly converges with fewer fine-tuning iterations. The Simpler-Env results further demonstrate the effectiveness of world-model-based post-training for efficient policy adaptation across diverse robotic setups.
While similar effects are observed in latent-action methods~\cite{ye2024latent,chen2024moto, gao2025adaworld}, our world model offers a simpler paradigm without latent actions, achieving better transferability.
\begin{table*}[htbp]
\centering
\caption{\textbf{Post-training enables data-efficient and training-efficient downstream policy learning.}}
\label{tab:post_efficiency}
\begin{subtable}[t]{0.48\textwidth}
\centering
\caption{\textbf{Data efficiency comparison.}}
\resizebox{\linewidth}{!}{%
\begin{tabular}{lcc}
\toprule
\textbf{Method} & \textbf{Data} & \textbf{CALVIN} \\
\midrule
RT-1~\cite{brohan2022rt} & 10\% & 0.34\\
MT-R3M~\cite{nair2022r3m} & 10\% & 0.61\\
HULC~\cite{mees2022matters}  & 10\% & 1.11\\
GR-1~\cite{wu2023unleashing} & 10\% & 2.00\\
RoboVLMS~\cite{li2024towards} & 10\% & 2.52\\
\midrule
\textbf{\methodname{}} (w/o post-train) & 10\% & 0.15 \\
\rowcolor[gray]{0.9}\textbf{\methodname{}} & 10\% & \textbf{3.19} \\
\bottomrule
\end{tabular}
}
\label{tab:data_efficiency}
\end{subtable}
\hfill
\begin{subtable}[t]{0.48\textwidth}
\centering
\caption{\textbf{Training efficiency comparison.}}
\resizebox{\linewidth}{!}{
\begin{tabular}{lccc}
\toprule
\multicolumn{4}{c}{\textbf{Fast convergence (CALVIN)}} \\
Training Iters & 2k & 4k & 8k  \\
\midrule
w/o post-train & 0.37 & 0.82 & 1.46  \\
\rowcolor[gray]{0.9}w/ post-train    &4.21 & 4.56 & 4.61    \\
\midrule
\multicolumn{4}{c}{\textbf{Fast adaptation (SimplerEnv-Bridge)}} \\
Method & Batch size & Iters & Success \\
\midrule
RoboVLMs~\cite{li2024towards} & 128   &50k & 37.5 \\
\rowcolor[gray]{0.9}\textbf{\methodname{}  }    & 128 &12k& 64.6 \\
\bottomrule
\end{tabular}
}
\label{tab:train_efficiency}
\end{subtable}
\vspace{-5mm}
\end{table*}

\begin{table*}[htbp]
    \centering
    \caption{\textbf{Ablation study on the visual prediction and historical context in policy learning.}}
    \begin{subtable}[t]{0.52\linewidth}
        \centering
        \caption{\textbf{Effectiveness of visual prediction.}}
        \resizebox{\linewidth}{!}{%
        \begin{tabular}{cccc}
            \toprule
            \textbf{Post-train} & \textbf{Visual prediction} & \textbf{CALVIN} & \textbf{LIBERO} \\
            \midrule
            \checkmark & & 4.61 & 94.2\\
            \rowcolor[gray]{0.9} & \checkmark & 4.42 & 88.7 \\
             & & 1.46 & 48.5 \\
            \bottomrule
        \end{tabular}
        }
        \label{tab:calvin_abl_visual}
    \end{subtable}
    \hfill
    \begin{subtable}[t]{0.46\linewidth}
        \centering
        \caption{\textbf{Effectiveness of history context.}}
        \resizebox{\linewidth}{!}{%
        \begin{tabular}{ccc}
        \toprule
        \multicolumn{2}{c}{\textbf{Observations}} & \textbf{Avg. Len $\uparrow$} \\
        \cmidrule(lr){1-2}
        \textbf{History Window} & \textbf{Current + History} & \\
        \midrule
        0 & 1 + 0 & 4.26 \\
        \rowcolor[gray]{0.9}10 & 1 + 1 & 4.61 \\
        10 & 1 + 2 & 4.43 \\
        20 & 1 + 2 & 4.47 \\
        \bottomrule
\end{tabular}
        }
        \label{tab:calvin_abl_history}
    \end{subtable}
\end{table*}

\vspace{-2mm}
\paragraph{Effectiveness of Visual Prediction.}
While post-training proves effective, it is also crucial that the model demonstrates strong performance without relying on it. As shown in Table~\ref{tab:calvin_abl_visual}, our findings indicate that, even without post-training, fine-tuning with visual loss supervision—leveraging the autoregressive nature of the model—naturally integrates world model learning into the policy learning process. This approach leads to a significant improvement in the model's performance.
\vspace{-2mm}
\paragraph{Effectiveness of History Context.}
History context—comprising past observations and actions—provides valuable guidance for robot planning. In this section, we investigate the appropriate length of the history window during the fine-tuning stage. As shown in Table~\ref{tab:calvin_abl_history}, our ablation study on the CALVIN benchmark examines the impact of varying history window lengths. Incorporating a history window significantly improves performance (from 4.26 to 4.61). However, extending the window beyond a certain length yields diminishing returns, suggesting that recent observations carry the most predictive value, consistent with the Markov property in sequential planning.

\subsection{Multimodal Capability}
\begin{figure}[tbp]
    \centering
    \includegraphics[width=\textwidth]{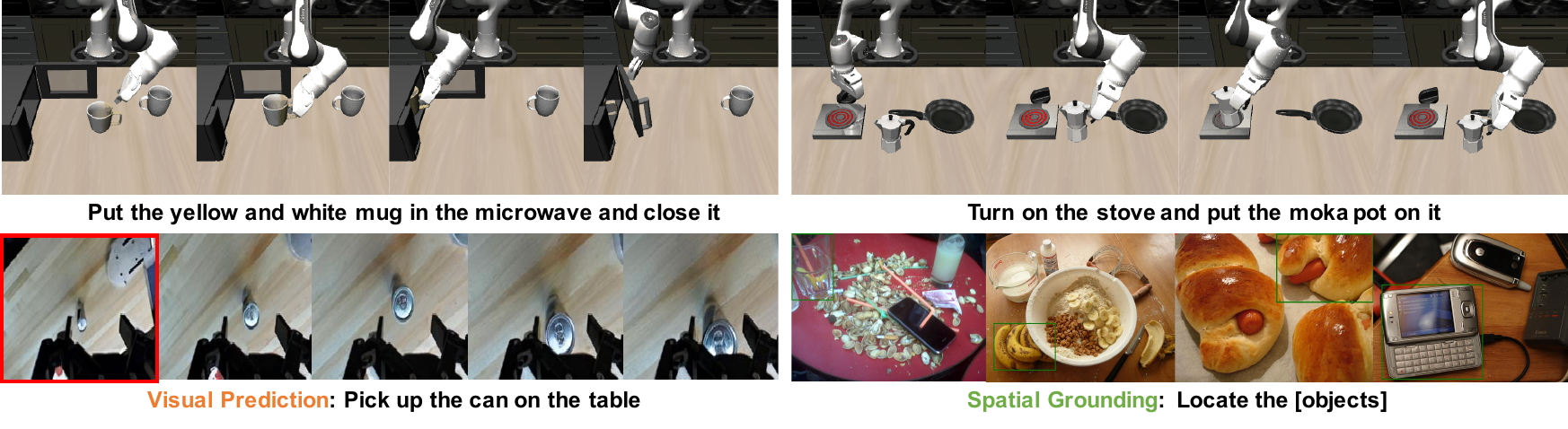}
    \vspace{-5mm}
    \caption{\textbf{Multimodal capabilities of \methodname{}}. Top: Action outputs for executing long-horizon tasks in the LIBERO benchmark. Bottom: Visual predictions and spatial grounding demonstrating the model's spatiotemporal understanding. The red box marks the current observation; green boxes indicate predicted object detections.}
    \label{fig:visual_demo}
    \vspace{-4mm}
\end{figure}

As illustrated in Figure~\ref{fig:visual_demo}, we qualitatively showcase the model’s ability to interleave multiple modalities—action, language, and vision—within a unified framework. This design enables policy learning for embodied control, spatial reasoning through language output, and future state prediction via visual output, highlighting the model’s capacity for generalizable multimodal understanding.

\subsection{Broader Applications}
\vspace{-1mm}
\paragraph{End-to-end Learning for Autonomous Driving.}
To further explore the potential of our method, we perform a preliminary transfer to the autonomous driving domain by finetuning the model on the NAVSIM benchmark. Notably, our method is a pure autoregressive, token-based framework, modeling the driving task as causal sequence prediction over discretized multimodal tokens. Despite using only front-view camera inputs—without relying on BEV representations or multi-sensor fusion—our model achieves powerful performance on the NAVSIM test set. Notably, the current performance is not pretrained on driving videos but is only fine-tuned on downstream policy benchmarks. These results highlight the strong potential of our method for broader real-world applications.
\begin{table}[htbp]
  \centering
    \caption{\textbf{Broader applications of \methodname{} for end-to-end autonomous driving on the NAVSIM.} MC: Multi Camera. L: LiDAR. FC: Front Camera.
    }
  \resizebox{0.88\linewidth}{!}{
  \begin{tabular}{lccccccc>{\columncolor{gray!20}}c}
    \toprule
    \textbf{Method} & \textbf{Model} & \textbf{Input} & \textbf{NC$\uparrow$} & \textbf{DAC$\uparrow$} & \textbf{EP$\uparrow$} & \textbf{TTC$\uparrow$} & \textbf{C$\uparrow$} & \textbf{PDMS$\uparrow$} \\
    \midrule
    \textcolor{gray}{Human} & \textcolor{gray}{--} & \textcolor{gray}{--} & \textcolor{gray}{100.0} & \textcolor{gray}{100.0} & \textcolor{gray}{87.5} & \textcolor{gray}{100.0} & \textcolor{gray}{99.9} & \textcolor{gray}{94.8} \\
    \midrule
    Ego Status MLP & -- & Ego State & 93.0 & 77.3 & 62.8 & 83.6 & 100.0 & 65.6 \\
    VADv2~\cite{vadv2}          & BEV-based & MC & 97.9 & 91.7 & 77.6 & 92.9 & 100.0 & 83.0 \\
    UniAD~\cite{uniad}          & BEV-based & MC & 97.8 & 91.9 & 78.8 & 92.9 & 100.0 & 83.4 \\
    Transfuser~\cite{Transfuser}     & BEV-based & MC\&L & 97.7 & 92.8 & 79.2 & 92.8 & 100.0 & 84.0 \\
    \midrule
    \textbf{\methodname{}} & Auto-regressive & FC & 96.9 & 91.1 & 76.8 & 91.7 & 96.7 & 81.7 \\
    \bottomrule
  \end{tabular}}
  \label{tab:driving_navsim}
  \vspace{-5mm}
\end{table}

\section{Conclusion}
In this paper, we present \methodname{}, a unified framework for vision–language–action modeling that bridges heterogeneous modalities through a shared token space and models them autoregressively. The proposed unified design facilitates deeper cross-modal integration and inherently supports flexible multimodal tasks. By leveraging a world model trained to capture dynamics and causality from videos, we observe significant improvements in downstream policy learning, both in terms of performance and efficiency. Extensive simulation experiments further demonstrate the model’s strong generalization ability, efficient policy learning, and broad applicability across diverse domains.
These findings highlight the great potential of our method as a new paradigm for vision–language–action modeling.
\vspace{-2mm}
\paragraph{Limitations and Future Work.}
Due to limited computational resources, our investigation into post-training scalability is still in its early stages. Nonetheless, initial results are promising and indicate potential for scaling to larger video datasets. Furthermore, while the unified multimodal framework exhibits strong capabilities in cross-modal learning, further research is needed to fully integrate it with reinforcement learning paradigms, enabling more robust and adaptive policy learning.

\clearpage
\bibliographystyle{plain}
\bibliography{main}
\clearpage

{
\small
}
%%%%%%%%%%%%%%%%%%%%%%%%%%%%%%%%%%%%%%%%%%%%%%%%%%%%%%%%%%%%

\appendix
\section*{Appendix}
\section{Implementation Details}
\paragraph{Post-training Stage}
We began by selecting several high-quality robotics datasets for post-training, as summarized in Table~\ref{tab:dataset_weights}. To account for differences in data collection frequencies across datasets, we applied dataset-specific frame sampling intervals to ensure that the temporal gap between keyframes is approximately one second. We further filtered out video sequences containing fewer than six frames, as well as those lacking corresponding text instructions.
Due to the large number of videos from the Kuka~\cite{kalashnikov2018scalable} dataset, we randomly retained 100k videos to prevent it from dominating the overall training data.
\begin{table}[htbp]
\centering
\small
\caption{\textbf{Post-training datasets}.}
\begin{tabular}{lccccc}
\toprule
\textbf{Dataset}& \textbf{Source} &\textbf{Data Type} & \textbf{Raw Videos} & \textbf{Used Videos} & \textbf{Interval} \\
\midrule
RT-1~\cite{brohan2022rt} & Real & Text, Video, Action  & 87212 & 84084& 3  \\
BridgeV2~\cite{walke2023bridgedata}& Real &Text, Video, Action  & 
60064 &28083 & 5\\
DROID~\cite{khazatsky2024droid}& Real &Text, Video, Action  &275997 & 145641& 15\\ 
Kuka~\cite{kalashnikov2018scalable} & Real &Text, Video, Action  & 580392& 100000& 3\\ 
TOTO~\cite{zhou2023train} &  Real & Text, Video, Action  &902 &899 & 20\\
Taco Play~\cite{rosete2023latent} & Real & Text, Video, Action  & 3242&3242 & 5\\
FMB~\cite{luo2023fmb}& Real & Text, Video, Action  &8611 &7876 & 5 \\
Berkeley autolab ur5~\cite{BerkeleyUR5Website} &Real &Text, Video, Action  & 896 & 896& 5 \\
% Robo net \\
VIOLA~\cite{zhu2023viola} &Real &Text, Video, Action &135 & 135& 15\\
Cmu Play Fusion~\cite{chen2023playfusion} &Real &Text, Video, Action&576 &576 & 10\\
Utaustin Mutex~\cite{shah2023mutex} &Real &Text, Video, Action& 1500& 1500& 10\\
\midrule
% 1x World Model~\cite{world_model_raw_data} & Real  & Video\\ \midrule
CALVIN~\cite{mees2022calvin} & Sim & Text, Video, Action  & 22966 & 22966&5 \\
LIBERO~\cite{liu2023libero} & Sim & Text, Video, Action  & 3386&3386 & 10\\ 
ManiSkill2~\cite{gu2023maniskill2} & Sim & Text, Video, Action  & 30213 &193273 &10\\
\midrule
SSV2~\cite{goyal2017something} & Real &Text, Video & 220847 &220847 & 1\\
\bottomrule
\end{tabular}
\label{tab:dataset_weights}
\end{table}

% experiments
For the experiments in Table~\ref{abl:post_train}, to ensure a fair comparison of different post-training strategies, all models are trained on the same dataset (excluding SSV2~\cite{goyal2017something}, which does not contain action annotations), with only the post-training strategy varied.
For the \emph{action prediction} task, we organize the input as $(T, I, A)$, where $T$ denotes the text instruction, $I$ the image observations, and $A$ the action sequence. During training, only the action tokens $A$ are supervised in the loss computation.
For the \emph{text-to-image} task, the input is organized as $(T, I)$, where $T$ denotes the input text and $I$ denotes the target image. During training, the loss is only computed on the vision tokens corresponding to $I$.
For the \emph{video prediction} task, the input is organized as $(I_1,...,I_t)$, where $I$ denotes the video frame. During training, the loss is computed on the vision tokens.
For the \emph{world model} task, the input is organized as $(T,I_1,...,I_t)$, where $T$ denotes the input text, $I$ denotes the video frame. During training, the loss is computed on the vision tokens.

% compute
During training, the observations are resized to 256$\times$256, using six frames as input, with the maximum sequence length set to 6400. We perform full-parameter training for 50k steps using 32 A100 GPUs (40GB), which takes approximately 4–5 days.

\paragraph{Simulation Finetuning}
The training setup is described in the main paper. We adopt full-parameter training, and for evaluation, we follow the testing protocols of OpenVLA~\cite{kim2024openvla} and RoboVLMs~\cite{li2024towards} across various benchmarks. By default, our model is trained using video-format sequences; however, it also supports fine-tuning with image-format sequences. In the ablation study evaluating the effect of visual prediction, when post-training is not applied, the visual token weight is set to 0.5 while the action token weight is set to 1.0, in order to maintain balance between the two modalities.

\paragraph{Real-robot Finetuning}
For real-world evaluation, we conduct experiments on the ALOHA platform, using images captured from three perspectives: \texttt{cam high}, \texttt{wrist left}, and \texttt{wrist right}. 
The real-robot is controlled using end-effector (EE) pose. All input images are resized to a resolution of 128$\times$128. The model outputs a 14-dimensional action vector. The action chunk size is set to 20. For each task, we train for 8k steps with a batch size of 256. The learning rate is set to $5\times 10^{-5}$, and all other settings remain consistent with the above.
We also leverage world model pretraining, using video-based post-training on a collected real-aloha dataset (Table~\ref{tab:task_counts}). Interestingly, this post-training provides substantial benefits even when transferring to real-robot execution.

\section{Real-Robot Experiments}
\begin{figure}[htbp]
    \centering
    \includegraphics[width=\textwidth]{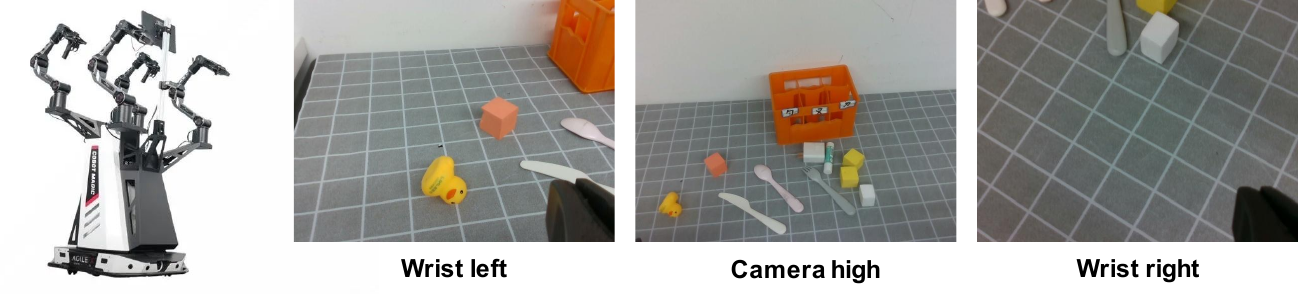}
    \caption{\textbf{Real-world setup of the AgileX Cobot Magic dual-arm robot.} The system is equipped with three RGB cameras for visual observation: one mounted on the left wrist, one on the right wrist, and one positioned above for a high-angle view.}
    \label{fig:songling}
\end{figure}
\subsection{ALOHA Experimental Setup}
\begin{figure}[htbp]
    \centering
    \includegraphics[width=\textwidth]{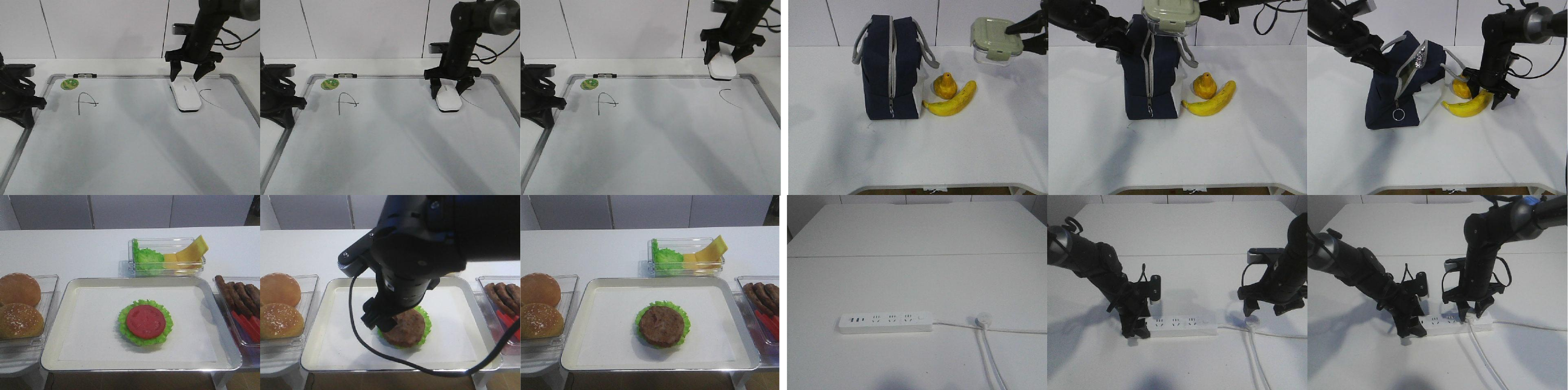}
    \caption{\textbf{Real-world task examples.} These include diverse tasks such as wiping a whiteboard, organizing tableware, making a burger, and plugging in a connector.}
    \label{fig:task}
\end{figure}

The robotic platform used in this paper is \textbf{AgileX Cobot Magic V2.0}, a dual-arm robot. As shown in Figure~\ref{fig:songling}, the robot is equipped with two arms and three camera views, enabling it to perform a variety of manipulation tasks. For example, Figure~\ref{fig:task} illustrates a range of manipulation tasks collected from real-world scenarios.

\paragraph{Real-World Task Collection}
Table~\ref{tab:task_counts} provides a summary of the real-world data collected from the physical robot, recorded at an actual frequency of 30 Hz. A total of 8 tasks were included, with each task collecting approximately 500 trajectories on average. During preprocessing, static frames at the beginning and end of each trajectory were filtered out.
\begin{table}[ht]
\centering
\caption{\textbf{Real-world task trajectories.}}
\resizebox{\textwidth}{!}{%
\begin{tabular}{c|c|c|c|c|c|c|c}
\toprule
\makecell{\textbf{Fold} \\ \textbf{Clothes}} &
\makecell{\textbf{Clear} \\ \textbf{Desk}} &
\makecell{\textbf{Store} \\ \textbf{Glasses}} &
\makecell{\textbf{Food} \\ \textbf{Packing}} &
\makecell{\textbf{Pour} \\ \textbf{Water}} &
\makecell{\textbf{Clean} \\ \textbf{Blackboard}} &
\makecell{\textbf{Insert} \\ \textbf{Plug}} &
\makecell{\textbf{Make} \\ \textbf{Hamburger}} \\
\midrule
528 & 500 & 500 & 500 & 496 & 500 & 500 & 640  \\
\bottomrule
\end{tabular}%
}

\label{tab:task_counts}
\end{table}

\paragraph{Data Processing}
To reduce redundancy and improve training efficiency, we select keyframes based on thresholding the changes in recorded action joint values. For each selected sequence, the action chunk is normalized by subtracting the joint values of the first frame.

\section{Autonomous Driving Experiments}
\paragraph{NAVSIM Setup}
The NAVSIM dataset~\cite{dauner2024navsim}, resampled from OpenScene to emphasize challenging scenarios, is currently one of the most established end-to-end evaluation benchmarks in the autonomous driving domain. The dataset is divided into two parts: Navtrain and Navtest, comprising 1,192 scenarios for training and validation, and 136 scenarios for testing.

For model training, the input images are resized to a resolution of 512$\times$288. We follow the standard training setup, using the current image frame and ego status to predict trajectories for the next 8 frames. Both the action and ego status are encoded using the fast tokenizer.
\clearpage
%%%%%%%%%%%%%%%%%%%%%%%%%%%%%%%%%%%%%%%%%%%%%%%%%%%%%%%%%%%%
% \input{secs/checklist}
\end{document}